\documentclass[journal]{IEEEtran}
\usepackage{color}
\usepackage{graphicx}
\usepackage{amssymb}
\usepackage{amsmath}
\usepackage{multirow}

\ifCLASSINFOpdf
\else
\fi
\begin{document}

\title{An Open-source End-to-End Logic Optimization Framework for Large-scale Boolean Network with Reinforcement Learning}

\author{Zhen~Li, Kaixiang~Zhu, Lingli~Wang,~\IEEEmembership{Member,~IEEE,}
        and~Xuegong~Zhou
        
}

\maketitle

\begin{abstract}
Large scale network,
\end{abstract}

\begin{IEEEkeywords}
Logic synthesis, Partition, Reinforcement learning
\end{IEEEkeywords}

%
\IEEEpeerreviewmaketitle

\section{Introduction}
\IEEEPARstart{L}{ogic} synthesis stands as a crucial process in Electronic Design Automation (EDA) tools for Integrated Circuits (ICs) design. Logic synthesis refers to the process producing a compact and fast gate-level implementation from a Register-Transfer Level (RTL) design description \cite{10.5555/541643}. The traditional process of logic synthesis includes translation, logic optimization, and technology mapping. The overall goal of logic synthesis is to find the best implementation of a boolean function, from the aspects of several performance metrics\cite {QY}.

At present, the popular academic synthesis tools optimize the circuit incrementally through a serious of predefined fixed synthesis transformations, which may not generate the best quality of results (QoR) for all possible logic circuits. To achieve better synthesis performance, synthesis scripts need to be customized manually for each input design. With the development of machine learning (ML), its integration with EDA has emerged as a burgeoning frontier in technology. For logic synthesis problems, For logic synthesis problems, reinforcement learning (RL) is leveraged to automatically and efficiently explore the the logic synthesis design space.

However, as the complexity and scale of integrated circuits continue to escalate, traditional methods of logic synthesis encounter formidable challenges. First, modern System-on-Chip (SoC) architectures are heterogeneous and composed by different Intellectual Property (IP)
blocks. The conventional one-size-fits-all approach to logic optimization falls short in addressing the unique requirements and characteristics of different regions within a large circuit. Second, the exponential large search space of the available logic transformation makes logic synthesis optimization for a large circuit extremely time-consuming. Therefore, it is desirable to partition a large heterogeneous circuit and individually optimize each portions. This approach can enhance the QoR of the circuit and improve the efficiency of logic synthesis optimization simultaneously.

In our work, we propose a novel methodology that leverages circuit partitioning and enhances logic optimization with RL,  aiming for efficient and effective optimization of large-scale Boolean networks.
The main contribution of this work are as follows:
\begin{itemize}
    \item A partitioning method for large-scale Boolean network circuit is proposed. Our experiments demonstrate that conducting logic synthesis optimization on the sub-circuits obtained through this partitioning method yields better QoR.
    \item A parallel RL-based logic synthesis flow exploration is conducted for the partitioned sub-circuits, greatly reducing runtime and significantly increasing scalability.
    
    {\color{red}\item     
    We demonstrate the capabilities of our proposed approach on the EPFL \cite{epfl}, OPDB \cite{OPDB} VTR \cite{vtr8} and Koios \cite{koios} benchmarks with the ASAP 7nm standard cell library \cite{ASAP7}.  We compare our work to logic optimization without circuit partitioning. We show that xxx outperforms previous techniques \cite{QY,DRiLLS,boils,LSOracle}}
    
\end{itemize}

The rest parts of this paper is organized as follows: Section II introduces the background information of logic synthesis and reinforcement learning. Section III describes the fine-grained partitioning of large circuits. In section IV, the end-to-end logic optimization framework for large-scale boolean network is proposed. In Section V, the performance results of the proposed framework are evaluated for several tasks and compared with existing works. Finally, Section VI summarizes this article.

\section{Preliminaries And Related Work}

\subsection{Boolean Logic Optimization}
Typically, a boolean network of the circuit is represented by a directed acyclic graph (DAG), wherein nodes representing logic gates and directed edges representing wires connecting the gates.  A well-known example of a Boolean network used in logic synthesis is the And-Inverter Graphs (AIGs), which is used by the state-of-art academic logci synthesisi tool ABC \cite{ref_abc}. AIGs consist of nodes representing logic AND functions and weighted edges indicating the inversions of the Boolean signals. Similarly, other forms of DAGs exist as well. MIGs were introduced in \cite{MIG} and XAGs were introduced in \cite{XAG} as a replacement for AIGs. Boolean logic optimization is typically performed by the algebraic minimization of a boolean network. The most effective algorithms that optimize the boolean network are based DAG-aware boolean transformations. DAG-aware boolean transformations have already been widely used in academic tools \cite{ref_abc}, such as \textit{rewrite}, \textit{refactor}, and \textit{re-substitute}. The existing ML-based exploration works are mainly implemented on AIGs. This is because AIGs-related logic optimization algorithms are the most comprehensive, providing a broader exploration space for logic synthesis optimization.


\subsection{Circuit Partitioning}
Circuit partitioning is the process of dividing a large and intricate circuit into smaller, more manageable clusters, usually with minimized inter-cluster connections. Due to the complexity of large-scale circuits, traditional optimization algorithms is time consuming for a single iteration over the entire system, rendering the optimization process inefficient. By partitioning the circuit into smaller portions, each segment can be optimized independently, significantly reducing the time required for each iteration. Partitioning plays an important roles in in modern EDA flows \cite{partitioning}. For example, during the physical design stage, partitioning is instrumental in floorplan\cite{floorplace}, placement \cite{place_ref}, clock tree synthesis \cite{cts_ref} and grouping instances within distinct power domains \cite{power}. 


\subsection{Reinforcement learning}
The advances in ML and specially RL open up new opportunities for logic optimizations. RL stands out due to its unique approach of learning to make decisions. It involves an agent that learns to achieve a goal in an uncertain, potentially complex environment. In RL, an agent learns from the consequences of its actions, rather than from being taught explicitly. It makes decisions, receives feedback from the environment in the form of rewards or penalties, and then adjusts its actions accordingly. By treating the optimization problem as an environment and the decisions regarding logic transformations as actions, an RL agent can be trained to navigate the vast search space of possible optimizations. 
RL has been implemented in logic optimizer to navigate the vast search space of possible optimization sequences to identify strategies that yield the best trade-offs between circuit performance metrics, such as power, area, and delay. Studies such as DRiLLS, ESE and BOiLS have demonstrated the feasibility and effectiveness of using RL in logic synthesis, paving the way for more efficient and automated EDA processes \cite{QY,DRiLLS,boils}.

\subsection{Prior works}

Previous works on logic synthesis for heterogeneous circuits mainly focus on representing different circuit portions with various data structures \cite{LSOracle,LSOracleDATE,LSOracleDAC}. 
Specifically, the work \cite{LSOracle,LSOracleDATE} proposed a logic synthesis framework, LSOracle, designed to select the most appropriate Boolean network data structure for different portions of logic. This includes And-Inverter Graphs (AIGs) and Majority-Inverter Graphs (MIGs). Building upon this, the work \cite{LSOracleDAC} further extends the framework by incorporating XOR-And Graphs (XAGs) into the Boolean network data structure used in the logic synthesis process. 
In \cite{LSOracle}, k-way hypergraph partitioning is employed during logic synthesis phase to break the original DAG into several partitions. Each of these partitions is represented with various data structures. 
{\color{red} However, our experimental findings suggest that the optimization impact of different Boolean network data structures is somewhat limited, with AIGs continuing to demonstrate superior efficiency as a Boolean network representation method.} 

The application of RL in logic synthesis, as explored in \cite{DRiLLS,boils}, adopts a holistic approach to circuit optimization. However, it has been observed that different segments of a large circuit could potentially derive greater benefits from tailored logic optimization scripts. Furthermore, the process of exploring and training to determine these optimization sequences for large circuits is notably time-consuming and demands extensive computational resources. Acknowledging these challenges, we propose an innovative approach that combines circuit partitioning with reinforcement learning logic optimization. This method aims to minimize the interdependencies between partitioned circuits while simultaneously enhancing runtime efficiency through parallel processing. The ultimate goal is to achieve an improved QoR.

\section{Partitioning Strategy}
\begin{figure}[!t] 
\centering
\includegraphics[width=3.2in]{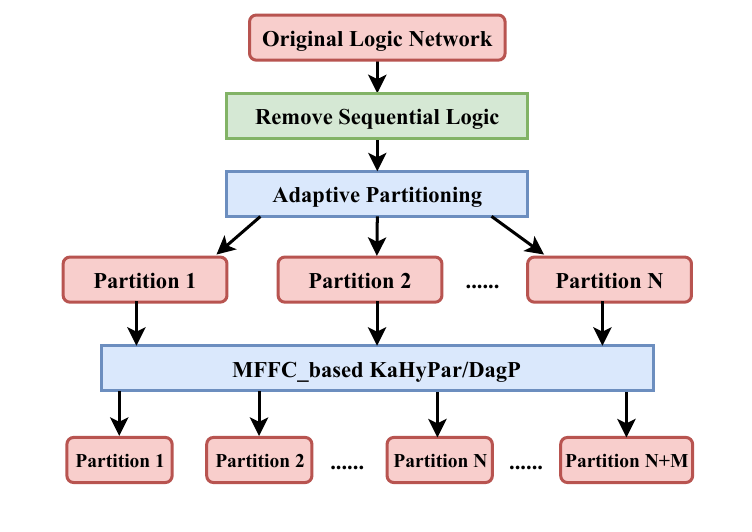}
\caption{PartitionFlow}
\label{fig:PartitionFlow}
\end{figure}

In the logic synthesis phase, the conventional wisdom suggests that circuit partitioning might compromise the quality of logic synthesis optimization, as partitioning could potentially degrade the effectiveness of mapping and logic optimization. However, we have discovered that partitioning based on the maximum fanout-free cone (MFFC) exhibits inherent suitability for logic synthesis optimization. Our experimental results indicate that not only does it not lead to a decline in optimization quality, but when combined {\color{red}with the optimization of logic synthesis sequences}, it actually enhances the QoR. During logic synthesis, partitioning a DAG based on MFFC enables increased independence among individual individual subgraphs. This minimizes the impact of logic optimization for one subgraph on others, thereby enhancing the efficiency of logic synthesis and ultimately improving the QoR. In our work, we integrated adaptive partitioning \cite{abcP} with MFFC-based k-way Hyper-Graph Partitioning (KaHyPar) \cite{KaHypar} and Directed Acyclic Graph Partitioning (DagP) \cite{dagP} to divide the logic network into subcircuits. To accelerate the overall logic synthesis optimization process, we limit the size of each subcircuits to no more than 10,000 nodes.  Fig.\ref{fig:PartitionFlow} presents a high-level view of the proposed partitioning approach.

\subsection{Adaptive Partitioning}

The adaptive partitioning approach in our study utilizes a cluster-based strategy, where the circuit is naturally segmented into independent clusters by excluding combinational logic's registers and macro-blocks, such as DSPs and BRAMs. This exclusion naturally divides the original netlist into independent clusters, particularly beneficial for large-scale pipelined circuits. These clusters, defined as sub-netlists with no common nodes, are formed without cutting any edges in the netlist, making the cluster-based strategy preferable. 
Clustering information is generated by traversing all nodes twice in reverse topological order, determining cluster membership. To ensure a balanced workload across clusters, all nodes are traversed twice in reverse topological order to determine their cluster membership, and the potential workload of each cluster is calculated by simulating the mapping process.
Given the clusters with their workload, the netlist can be naturally partitioned into several workload-balanced sub-netlists without cutting any edges.

While most large-scale circuits exhibit natural division by registers and macro-blocks, some circuits may possess a single large, interconnected cluster impeding cluster-based partitioning. For such cases, the open-source partitioning tool named Metis is deployed.  Metis partitions the netlist by cutting edges with minimized weight sums, carefully avoiding critical paths by assigning them higher weights to maintain delay constraints.
Edges cut by Metis may require additional handling to maintain sub-netlist integrity. Additional PI and PO nodes are generated for each cut edge, ensuring connectivity across partitions. During sub-netlist merging, these additional nodes are omitted, and internal logic nodes are connected directly, preserving the structural integrity of the mapped netlist.

\subsection{MFFC-Based KaHyPar and DagP}

Following the implementation of the adaptive partitioning strategy outlined in the preceding section, we observed that certain sub-netlists continue to exhibit large sizes, thereby rendering the subsequent logic optimization processes both time-consuming and challenging. In response to this issue, we have opted to utilize a MFFC-based partitioning approach to effectively manage instances of oversized sub-netlists.

The MFFC decomposition technique was first proposed in 1993 for combination circuits \cite{MFFC_93}.
It offers a method to analyze and partition circuits based on their logical structure and signal flow. In this context, a fanout-free cone (FFC) at a given node \(v\) is a subset of the circuit that includes \(v\) and all its predecessors, ensuring that any signal path to \(v\) remains within this subset. The MFFC of a node \(v\), then, represents the largest FFC that includes \(v\) and adheres to the condition that if any non-primary input (PI) node \(w\) has all its outputs within the MFFC, then \(w\) is also included in the \(MFFC_v\). This technique is notable for its ability to uniquely identify closely related gates within a circuit, as any gate within an MFFC contributes directly to the output of the MFFC's root node. 

Key properties of MFFC decomposition include:
\begin{itemize}
    \item  If $w \in MFFC_v$, then $ MFFC_w \subset  MFFC_v$.
    \item  Any two MFFCs are either completely separate or one encompasses the other.
\end{itemize}

\begin{figure}[!t] 
\centering
\includegraphics[width=3.6in]{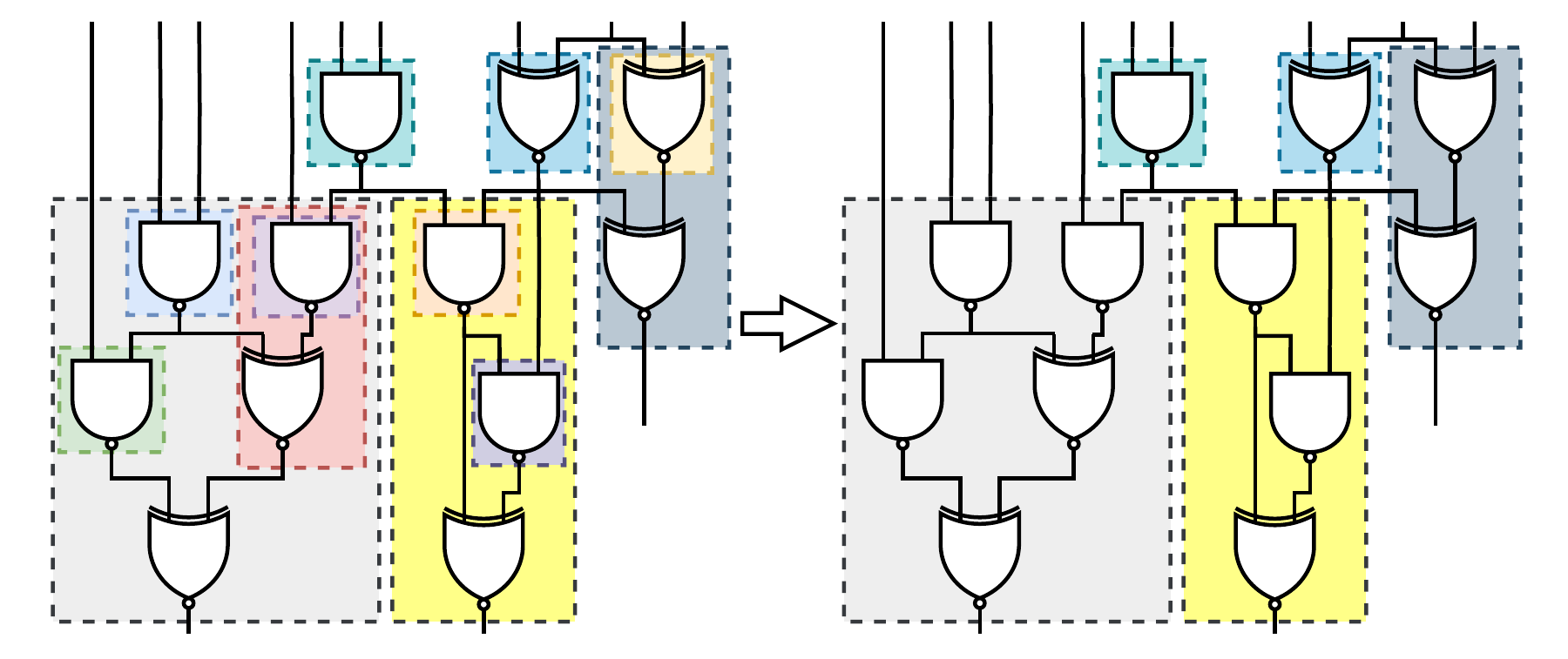}
\caption{ MFFC decomposition}
\label{fig:mffc}
\end{figure}

Based on these characteristics, a combinational circuit N can be uniquely decomposed into disjoint MFFCs by: (i) selecting a primary output node v and computing its MFFC, (ii) redefining N to exclude MFFC and treating nodes with fanouts to MFFC as new POs, and (iii) repeating the process recursively. Figure \ref{fig:mffc} illustrates an example of the MFFC decomposition. 
MFFC decomposition ensure the logic independency of the partitioned subgraphs, thereby not only facilitating efficient optimization of logic synthesis but also minimizing the impact of optimizing one subgraph on others. This approach significantly enhances the overall efficiency of the logic synthesis process.


After adaptive partitioning, we continue to divide the oversized sub-nwelists into smaller portions. The MFFCs for every node in the target Boolean network is traversed and constructed. Then the original graph in compressed into a new generated graph, where each identified mffc is conceptualized as a node and dependencies among MFFCs are structured as edges. 


To effectively partition this newly generated graph, we employ either KaHyPar or DagP, aiming primarily to minimize the disruption of edges. KaHyPar, an open-source library, leverages a multilevel approach for Hyper-Graph Partitioning (HGP), utilizing a progressively refining strategy to efficiently handle large datasets and deliver high-quality partitions. It excels in dividing hypergraphs into k distinct segments, with the goal of reducing the connector size between partitions, showcasing its innovative approach to partitioning.
Conversely, DagP adopts a multilevel methodology tailored specifically for HGP, with a particular focus on partitioning DAGs. It capitalizes on the topological ordering inherent in DAGs to boost parallel computation capabilities. By minimizing inter-partition dependencies, DagP enhances task execution within parallel computing environments, thereby reducing the latency caused by data transfers between partitions.

Employing these strategies significantly reduces dependencies between partitions, thereby improving the efficiency of parallel processing and facilitating more efficient management of large-scale computational tasks.  Additionally, this approach mitigates the mutual interference of logic synthesis optimization across different partitions, thereby enhancing the QoR of the circuit after merging.

\section{Proposed Logic Optimization Framework}

\begin{figure}[!t] 
\centering
\includegraphics[width=3.2in]{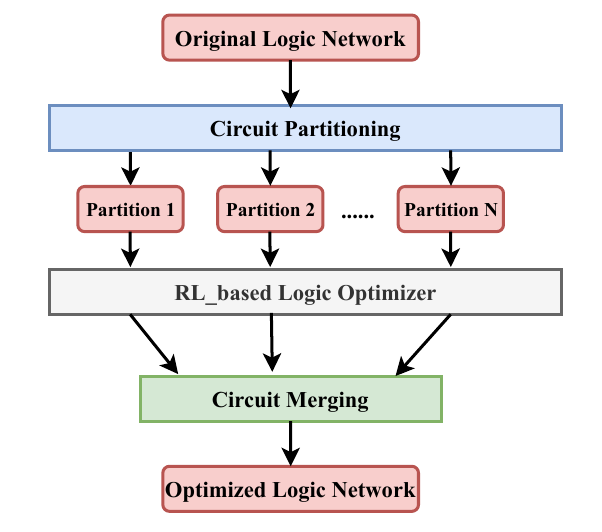}
\caption{Optimize Framework}
\label{fig:OptimizeFramework}
\end{figure}

Figure \ref{fig:OptimizeFramework} illustrates the logic optimization framework put forth in this study. The methodology commences with the $Original\ Logic\ Network$, establishing the baseline for the optimization. The foremost phase, $Circuit\ Partitioning$, segments the network into multiple partitions, each containing fewer than 10,000 nodes. This granular division is crafted to simplify the network's complexity, thereby enabling detailed optimization and expediting the process through parallelization. 

Subsequent to the partitioning, the $RL\_based\ Logic$ $Optimizer$ ESE from \cite{QY} embarks on refining each segment. Grounded in RL, this optimizer meticulously adjusts the logic optimizations sequences within each partition. 
Upon the culmination of individual optimizations, the partitions converge during the $Circuit\ Merging$ phase. Here, the optimized partitions are amalgamated to reconstruct the enhanced logic network. This crucial phase reintroduces the previously removed sequential elements, safeguarding the integrity of the network's functionality and performance. The framework thus ensures that isolated optimizations confluence seamlessly, maintaining the overarching efficacy and coherence of the original network.

\section{Experimental Results}

\subsection{Experimental Setup}

\begin{figure}[!t] 
\centering
\includegraphics[width=3.5in]{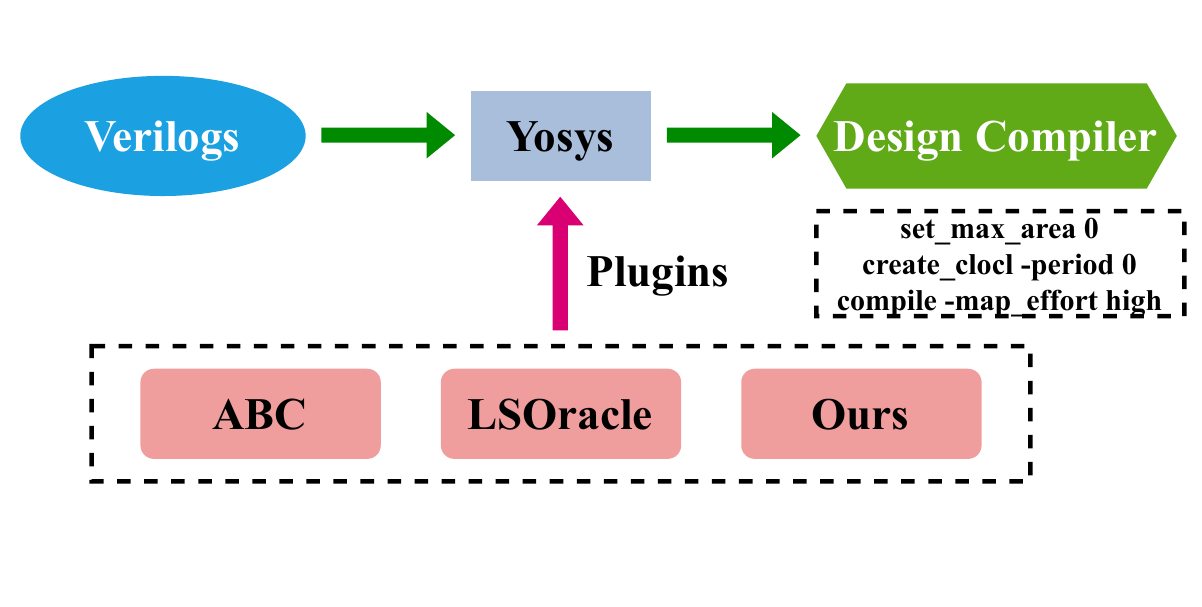}
\caption{ExperimentFlow}
\label{fig:ExperimentFlow}
\end{figure}

Our experimental flow is illustrated in Figure \ref{fig:ExperimentFlow}. We employ plugins to optimize the pure combinational logic within the circuit from the input Verilog files within the circuit using ABC, LSOracle, and our proposed method, respectively. Subsequently, the optimized Verilog is evaluated using Synopsys DesignCompiler, version s-2021.06-sp5. To minimize circuit optimization during mapping, we opted for the $compile$ command instead of $compile\_ultra$. Additionally, we set the clock to 0 to maximize the  maximize the maximum frequency.
Our experiment evaluated re sults from over 150 benchmarks sourced from EPFL \cite{epfl}, OPDB \cite{OPDB}, VTR \cite{vtr8}, and Koios \cite{koios}, excluding a few diminutive examples. These small cases were omitted due to the negligible optimization impact attributed to their size. Furthermore, one large benchmark named $hyp$ was removed, as it cannot be synthesized using DC within a week. The benchmarks were tested using the ASAP 7nm technology library \cite{ASAP7}.

We conducted a comparison between our work and that of DRiLLS \cite{DRiLLS}, BOiLS \cite{boils} and LSOracle\cite{LSOracle}. Our approach uses 200 cores to perform parallel optimization on the partitioned circuits for a duration of 2 hours. To ensure a fair comparison, for the single-core optimization works of BOiLS and DRiLLS, we set the optimization duration using the following formula: \( \lceil \frac{\text{number of parts}}{200} \times 2h \rceil \). Regarding the time taken to partition the circuits, for the largest circuit in our benchmark (comprising $10^6$ nodes), it required ten hours. We then synthesize the optimized Verilog files using  DC and generated reports on area, delay, and ADP (area delay product).

\subsection{End-to-end Results}

Our experimental results are presented in Table \ref{tab:comparison}.
The leftmost column indicates the size of the circuits, ranging from the largest 10, 20, 30, 40, to all benchmarks.
The second column delineates the metrics for comparison, including Area, Delay, and ADP (Area Delay Product), while the third to sixth columns display the optimization outcomes from LSOracle, BOiLS, DRiLLS, and our work, respectively. The baseline for comparison is the $resyn2$ command within ABC. 
As evident from the table, DRiLLS demonstrates the best optimization results for the largest 10 benchmarks, primarily focusing on delay improvement. However, for the subsequent groups ranging from the largest 20 to all benchmarks, our optimization method achieves superior enhancements in both delay and area, yielding an approximately 5\% improvement in ADP. The comparative analysis also reveals that LSOracle exhibits the least optimization benefits, with performance even falling short of ABC's $resyn2$, showing a negative optimization effect. We speculate conjecture that the reason behind this is that AIG remains the most efficient DAG representation to date. Consequently, partitioning circuits into different parts using various data structures for representation does not yield  performance improvements.

\begin{table}[]
\caption{Comparison with LSOracle, BOiLS, DRiLLS, and our approach.}
\label{tab:comparison}
\resizebox{0.5\textwidth}{!}{%
\begin{tabular}{c|ccccc}
\hline
\textbf{Benchmark Size}                          & \textbf{Metric}& \textbf{LSOracle}\cite{LSOracle} & \textbf{BOiLS}\cite{boils} & \textbf{DRiLLS}\cite{DRiLLS} & \textbf{Ours} \\ \hline
\multicolumn{1}{c|}{\multirow{3}{*}{Largest 10}} & Area  & 1.57\%   &-10.69\%&-12.00\%&-11.139\%\\
                                                 & Delay & 12.10\%  &0.69\% &-2.52\% &-0.23\%\\
                                                 & ADP   & 14.35\%  &-10.45\%&-14.11\%&-11.52\%\\ \hline
\multirow{3}{*}{Largest 20}                      & Area  & 1.29\%   &-5.80\%&-6.58\% &-6.55\%\\
                                                 & Delay & 13.61\%  &-0.69\%&-1.74\% &-1.78\%\\
                                                 & ADP   & 15.03\%  &-6.72\%&-8.11\% &-8.37\%\\ \hline
\multirow{3}{*}{Largest 30}                      & Area  & 5.03\%   &-5.78\%&-6.19\% &-4.87\%\\
                                                 & Delay & 11.12\%  &-0.08\%&0.09\%  &-0.82\%\\
                                                 & ADP   & 17.55\%  &-5.95\%&-5.88\% &-5.73\%\\ \hline
\multirow{3}{*}{Largest 40}                      & Area  & 4.10\%   &-5.04\%&-5.47\% &-5.50\%\\
                                                 & Delay & 8.90\%   &-0.55\%&-0.29\% &-1.49\%\\
                                                 & ADP   & 14.08\%  &-5.41\%&-5.46\% &-6.75\%\\ \hline
\multirow{3}{*}{All}                             & Area  & 2.08\%   &-3.28\%&-3.61\% &-3.85\%\\
                                                 & Delay & 3.81\%   &-0.64\%&-0.30\% &-1.70\%\\
                                                 & ADP   & 6.16\%   &-3.68\%&-3.73\% &-5.17\%\\ \hline
\end{tabular}
}
\end{table}

\section{Conclusion}

%



\section*{Acknowledgment}

The authors would like to thank...

\ifCLASSOPTIONcaptionsoff
  \newpage
\fi



%
\bibliographystyle{IEEEtran}
\bibliography{LogicSynth}
%








\end{document}